\documentclass[letterpaper, 9.5 pt, journal, twoside]{IEEEtran}

\usepackage{stackengine}
\usepackage{mathtools}
\usepackage{hyperref}
\usepackage{ragged2e}
\usepackage[english]{babel}
\usepackage{amssymb}

\usepackage{physics}
\usepackage{todonotes}

\usepackage{graphicx}
\ifCLASSINFOpdf
\else
\fi
\usepackage{amsmath}
\usepackage{amsfonts}
\usepackage[inline]{enumitem}
\usepackage{booktabs}

\begin{document}
%

\title{Autonomous Intraluminal Navigation of a Soft Robot using Deep-Learning-based Visual Servoing}

%
\author{Jorge F. Lazo$ \dagger^{1,2}$, 
Chun-Feng Lai$ \dagger^{1,3}$, 
Sara Moccia$^{4,5}$,
Benoit Rosa$^{2}$,
Michele Catellani$^{2}$,
Michel de Mathelin$^{6}$,\\
Giancarlo Ferrigno$^{1}$,
Paul Breedveld$^{3}$,
Jenny Dankelman$^{3}$,
and Elena De Momi$^{1}$
\thanks{$\dagger$ The authors contributed equally to this work.}
\thanks{$^{1}$ Jorge F. Lazo, Chun-Feng Lai, Giancarlo Ferrigno and Elena De Momi are with the Department of Electronics, Information and Bioengineering, Politecnico di Milano, Milan, Italy
}%
\thanks{$^{2} $Jorge F. Lazo, Benoit Rosa and Michel de Mathelin are with ICube, UMR 7357, CNRS-Université de Strasbourg, Strasbourg, France }%
\thanks{$^{3} $Chun-Feng Lai, Paul Breedveld and Jenny Dankelman are with Delft University of Technology, Delft, Netherlands 
}%
\thanks{$^{4,5}$Sara Moccia is with The BioRobotics Institute, Scuola Superiore Sant’Anna, Pisa, Italy and the Department of Excellence in Robotics and AI, Scuola Superiore Sant’Anna, Pisa, Italy
}
\thanks{$^{6}$Michele Catellani is with the European Institute of Oncology (IRCCS), Milan, Italy
}
\thanks{
This work was supported by the ATLAS project. This project has received funding from the European Union’s Horizon 2020 research and innovation programme under the Marie Skłodowska-Curie grant agreement No 813782.
This work was also partially supported by French State Funds managed by the Agence Nationale de la Recherche (ANR) through the Investissements d’Avenir Program under Grant ANR-11-LABX-0004 (Labex CAMI) and Grant ANR-10-IAHU-02 (IHU-Strasbourg).}
}

\maketitle
\begin{abstract}
Navigation inside luminal organs is an arduous task that requires non-intuitive coordination between the movement of the operator's hand and the information obtained from the endoscopic video.  
The development of tools to automate certain tasks could alleviate the physical and mental load of doctors during interventions, allowing them to focus on diagnosis and decision-making tasks. 
In this paper, we present a synergic solution for intraluminal navigation consisting of a 3D printed endoscopic soft robot that can move safely inside luminal structures. 
Visual servoing, based on Convolutional Neural Networks (CNNs) is used to achieve the autonomous navigation task. 
The CNN is trained with phantoms and in-vivo data to segment the lumen, and a model-less approach is presented to control the movement in constrained environments.
The proposed robot is validated in anatomical phantoms in different path configurations. 
We analyze the movement of the robot using different metrics such as task completion time, smoothness, error in the steady-state, mean and maximum error. 
We show that our method is suitable to navigate safely in hollow environments and conditions which are different than the ones the network was originally trained on.
\end{abstract}

\IEEEpeerreviewmaketitle

\section{Introduction}
Minimally invasive interventions present a wide range of benefits such as a smaller risk of wound infections, less damage to healthy tissue, and shorter postoperative hospital care. 
These types of medical procedures take advantage of already existing orifices or make use of small incisions, to reach the organs which need intervention.
In these scenarios, navigation inside narrow luminal organs such as the ureter, colon, or larynx could turn into a complex task, especially for less experienced operators~\cite{fisher2011complications}. 
Therefore, comprehensive training is required to master these techniques. 

Current constraints in endoluminal navigation can be considered three-fold. 
First, there are mechanical design limitations such as dimensions, steerability, and dexterity of the instruments~\cite{da2020challenges}. 
Second, image-related limitations, such as low image quality, the presence of artifacts, debris among others, can compromise procedures~\cite{ali2020objective}.
Finally, the coordination between the hand movement and the matching with the endoscopic image scenario is far from intuitive and could lead to hand-eye coordination problems~\cite{dankelman2011current}.

The necessity of performing intraluminal navigation in safer and more efficient ways, which can reduce possible complications such as tools colliding with tissues, mucosal abrasion, or minor perforations~\cite{de2006handling}, has led to a fast improvement of different models of Minimally Invasive Robotic Intervention (MIRI) systems~\cite{bergeles2013passive}.
Recently different levels of autonomy have been tested in a few prototypes~\cite{attanasio2021autonomy, boehler2020realiti}.
Visual servoing has been proposed to control different types of soft robots based on different actuation mechanisms. 
In the case of tendon-driven approaches, Wang et al. propose an adaptive visual servoing controller where the size of the manipulators is not required~\cite{wang2016visual}. The model was tested in open and confined space using a ring-shaped object to simulate a physical restriction. 
More recently, Lai et al. introduce a vision-based approach to control a soft robot manipulator composed of continuum segments of cable-driven mechanisms~\cite{lai2020toward}.
For Concentric Tube Robots (CTR), several studies have been conducted. 
Wu et al. propose a visual servoing approach based on tracking a laser target. This method does not require any previous knowledge on the kinematics model of the robot, just an initial estimation and a constant update of the Jacobian of the robot~\cite{wu2015model}.
Girerd et al. present a CTR that can navigate through origami tubular structures using a combination of a visual Simultaneous Localization And Mapping (SLAM) approach and a virtual repulsive force produced by the cloud points detected by the SLAM algorithm~\cite{girerd2020automatic}. 
%
Visual servoing has also been implemented in pneumatically driven robots.
Fang et al. use a pneumatically-driven 3-chamber robot and propose an eye-in-hand servoing method that incorporates a machine learning-based technique to estimate the inverse kinematic model without any prior knowledge about the model of the robot~\cite{fang2019vision}. 
The work presented by Wang et al.~\cite{wang2020eye} combines the use of template matching algorithms and Fiber Bragg Grating (FBG) sensors to achieve a more accurate tracking for visual servoing.

Zhang et al. developed a prototype of an endoscopic robot for the task of navigation in a colonoscopy scenario~\cite{zhang2020enabling}. Their prototype consists of biopsy forceps and an auto-feeding mechanism. Control was achieved using a workspace model to estimate the tool position. 
Martin et al.~\cite{martin2020enabling} presented a magnetic endoscope that can achieve different levels of autonomy as defined by Yang et al. in~\cite{yang2017medical}. Vision-based navigation is accomplished using a direction vector acquisition method. 
In the same clinical scenario, Prendergast et al.~\cite{prendergast2020real} introduce an autonomous navigation strategy using a finite state machine region estimation approach. 

Even though the approaches mentioned above are effective in specific scenarios, most of them are still based on the extraction of user-defined visual features or the use of extra sensing devices, which might make them prone to fail in scenarios with large variations in images conditions. 
Convolutional Neural Networks (CNNs) on the other hand tend to generalize better when they are trained in a large enough amount of data~\cite{ali2020objective}.
In this regard, we propose a CNN based on a model previously validated on patient image data~\cite{lazo2021using}, adapting it to a lighter version to be implemented in a robotic device. 
%

To address the current obstacles of intraluminal navigation, in this paper we propose an integrated solution which comprises: 
1) The implementation of a 3D printed flexible robot which allows fast prototyping, and simplicity in terms of scalability. 
2) A lumen-center detection system based on a CNN which can handle changing scenarios and variable image conditions,   
3) the synergic integration of the previous modules using a visual servoing control strategy to achieve autonomous navigation in narrow luminal scenarios.

We show the robustness of our approach by testing the navigation capabilities of the robot in different scenarios and phantoms which were not used to train the CNN. 
To the best of our knowledge, this is the first for autonomous intraluminal navigation MIRI system based on a CNN.

The main contributions of this work are: 

\begin{itemize}
    \item Effective integration of a 3D printed cable-driven flexible robotic endoscope with a model-less visual servoing system based on CNNs. 
    \item Validation of the proposed model-less control approach to bring the robot to the center of the lumen regardless of its initial position. 
    \item Demonstration of the capabilities of the robot to autonomously find the center of the lumen and  safely navigate through different intraluminal scenarios and paths which were not previously seen by the robot. 
\end{itemize}
  
An overview of our approach is provided in the supplementary video.

\begin{figure*}[tbp]
    \centering
    \includegraphics[width=0.75\textwidth]{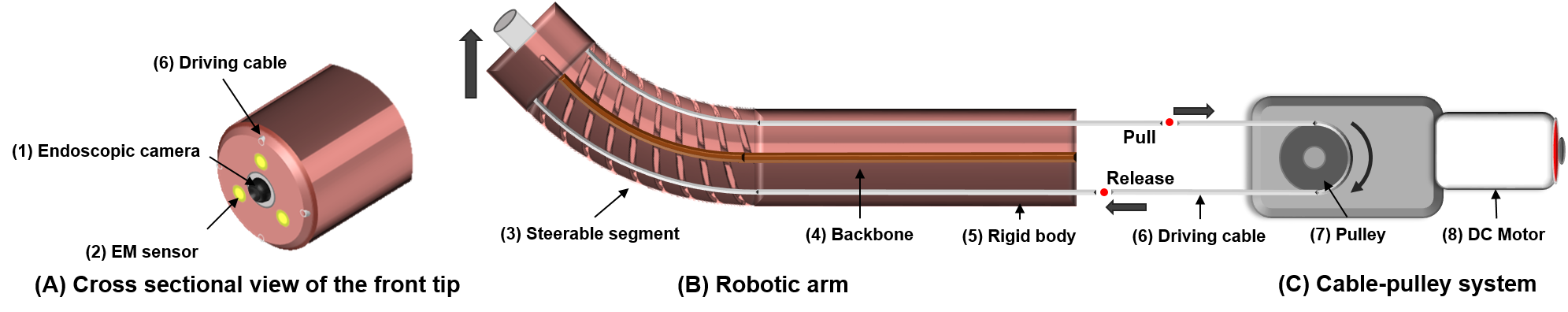}
    \caption{Illustration of the soft robotic mechanism. 
    A) 
    1) Endoscopic camera is placed in the center pointing to the front, 
    2) EM sensors illustrated in yellow were installed.
    B) Robotic arm, consists of three parts: 
    3) Steerable segment. The driving cables run through the arm and are glued onto the front tip, 
    5) Rigid body, 
    4) Backbone. and in
    C) Cable pulley system, 
    6) Driving cable was antagonistically arranged on the pulley. When the motor rotates, the driving cable will be pull and released at the same time;
    7) Pulley and, 
    8) DC motor with gearbox and encoder.
    }
    \label{fig:robot_CAD}
\end{figure*}

\section{System Overview}
\label{section:method}
\subsection{Robotic Platform}
To test the proposed visual servoing approach, a soft robotic endoscope prototype is manufactured based on the design of the non-assembly 3D-printed mechanism HelicoFlex~\cite{culmone2020exploring}.
The total length of the flexible robotic arm was set to 70 mm and and the outer diameter was set to 10 mm.
The prototype contains four channels with an inner diameter of 2.3 mm to allow sensors and cables from the endoscopic camera to pass through.
In the distal end of the prototype, a cup was designed for the installation of the endoscope camera module and Electromagnetic (EM) tracking sensors. 
It is worth noting that the diameter of robotic arm was manufactured at a 1:1 scale to a colonoscope, 2:1 scale to a cystoscope and 3:1 scale to a ureteroscope. 
While a smaller robotic arm could easily be manufactured thanks to the 3D printing technology used in our design, we chose a larger scale in this paper in order to easily install the EM sensors and the camera.
From the distal end of the steerable segment, four driving cables were inserted through the cable guides as illustrated in Fig.~\ref{fig:robot_CAD}. 
These cables pass through the rigid segment towards the actuation parts. 
The cables move freely within the cable guides and are glued to the tip of the prototype. 
Between the steerable segment and the actuation module, there is a rigid body with a total length of 58~mm. 
The prototype was 3D printed using digital light processing technology with a 3D printer (Perfactory 4 Mini XL, EnvisionTec Gmbh) and R5 epoxy photopolymer resin (EnvisionTec GmbH, Gladbeck, Germany). The minimum thickness of the printing layer height by the 3D printer was 0.02~mm, and the x-y accuracy on a printing layer was 0.03~mm.

\begin{figure}[tbp]
    \centering
    \includegraphics[width=0.48\textwidth]{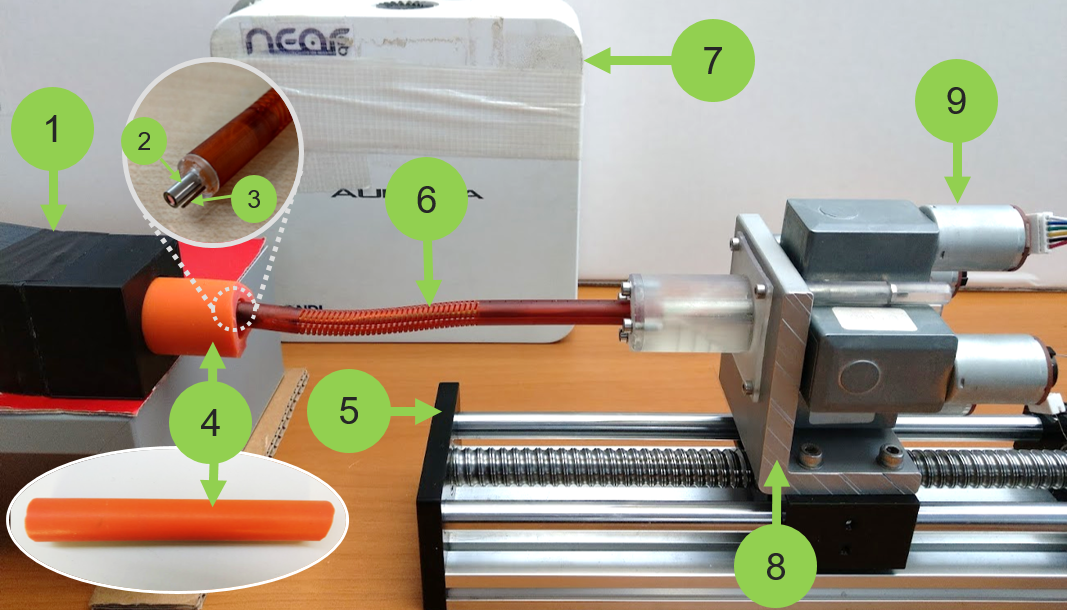}
    \caption{Assembly of the actuation platform for the soft robotic endoscope and the experimental set-up for the system validation:
    1) 3D printed mold to fix the curve of the lumen phantom;
    2) Endoscopic camera;
    3) Electromagnetic tracking sensors on the robot tip:
    4) Soft anatomical phantom;
    5) Linear stage;
    6) Soft robotic arm
    7) Electromagnetic field generator;
    8) Linear actuation module;
    9) DC motors
    }
    \label{fig:robot_setup_up}
\end{figure}

The actuation module of the soft robotic endoscope prototype, as shown in Fig.~\ref{fig:robot_setup_up}, consists of one linear stage and two DC motors (JGY370-30RPM, Walfront) with self-locked worm gearboxes. The DC motor has maximum 30 RPM and a nominal torque 7.4 Kg-cm.
Two cables on the opposite side of the flexible robot shaft were fixed antagonistically on one pulley and then connected to one DC motor (Fig.~\ref{fig:robot_CAD}).
This means that each motor pulley system is responsible for two steering cables and controlling one Degree of Freedom (DOF) of the soft robotic endoscope prototype.  
The prototype together with its motor pulley system is fixed onto a linear stage with a 200 mm (EBX1204-200, Garosa) stroke and a stepper motor (23SSM2440-EC1000, ACT MOTOR) to provide back and forth movement. 
In total, there are three controllable DOFs. 

\begin{figure}[tbp]
    \centering
    \includegraphics[width=0.49\textwidth]{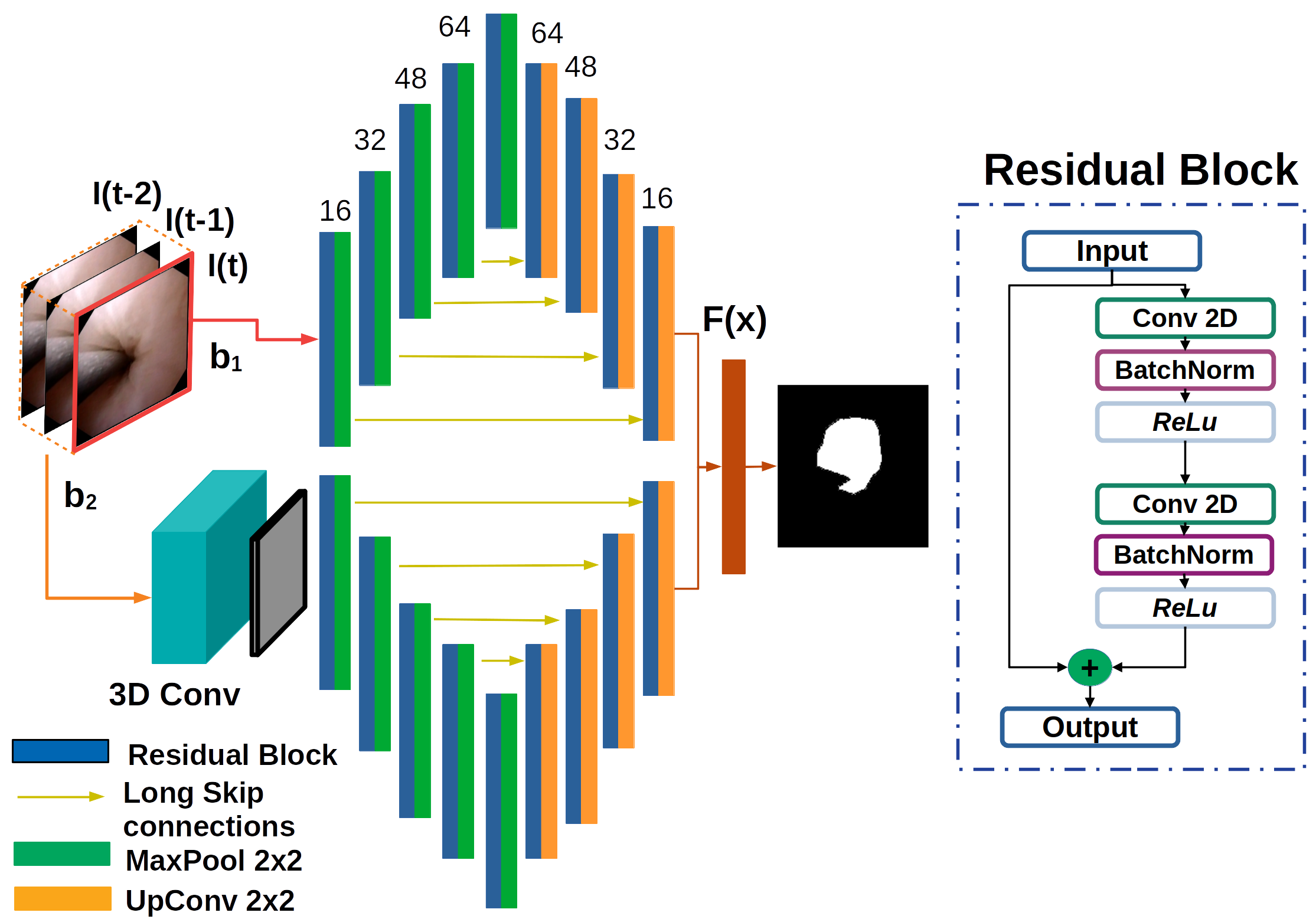}
    \caption{Architecture used for lumen segmentation. 
    The CNN is composed of two branches, both of them are composed by residual blocks. Branch $b_1$ process the information of the current frame, while while $b_2$ considers the information of the current, and three previous frames $I(t)$. The final output combines the predictions of both branches in the last layer using the ensemble function $F(x)$}
    \label{fig:cnn_architecture_used}
\end{figure}
\subsection{Lumen Center Detection}

The lumen center detection module consists of two steps, a lumen segmentation stage and a center detection algorithm. 
The first part, the lumen segmentation step, consists of an ensemble of CNNs based on~\cite{lazo2021using} and depicted in Fig.~\ref{fig:cnn_architecture_used}. 
To make the network resistant against temporal image variations, such as changes in the illumination, or the sudden appearance of image artifacts, the segmentation network takes as input three consecutive frames: $\lbrace I(t-2)$, $I(t-1)$, $I(t) \rbrace$ where $I(t)$ denotes the frame at moment $t$, and $t-1,  ..., t-n$ corresponds to the ordered previous frames. 
The input is processed by two parallel branches ($b_1$ and $b_2$). One of the branches processes the continuous blocks
while the other only handle individual frames $I(t)$. 

Residual blocks (dashed square Fig.~\ref{fig:cnn_architecture_used}) are the core units of both branches $b_1$ and $b_2$. They bifurcate internally in two branches, one of them is composed by two consecutive sets of convolutional layers, followed by a batch normalization layer using Rectified Linear Unit $ReLu$ as activation function. 
The final output of the residual branch is added to the identity input in the residual block. 
MaxPooling layers are used  for dimension reduction in the case of the encoding path, and 2D up-sampling layers for the decoding path, before connecting this layers to the subsequent residual block.  

The final output of the CNN is obtained using an ensemble function $F(x)$ followed by a sigmoid activation function. 
The ensemble function is defined as: 
\begin{equation}
    F(x) = \frac{1}{N} \sum^N_n x_n
\end{equation}
where $x_n$ corresponds to the individual outputs of the $b_n$ branches and N is equal to the number of the branches in the network.
%
The CNN is trained to minimize the loss function based on the Dice Similarity Coefficient Loss defined as:
\begin{equation}
    L_{DSC}= 1 - \frac{2TP}{2TP + FN + FP}
\end{equation}
where $TP$ (True Positives) corresponds to the number of pixels that are correctly segmented, $FN$ (False Negatives) is the number of pixels which are classified to be part of the lumen but actually they are not, and $FP$ (False Positives) refers to the pixels miss-classified as lumen. 

The center detection step is performed to determine the position of the target $\vb*{p} = (p_x, p_y)$ that the robot should follow. 
Considering the predicted mask $M$ obtained with the CNN, the moments of it can be obtained as:
\begin{equation}
    m_{i,j}= \sum_{u} \sum_{w} M(u,w) \cdot u^i \cdot w^j
\end{equation}
with $m_{i,j}$ the image moments and  $M(u,w)$ the pixels corresponding to the segmented area. The $p_x$, $p_y$ coordinates can be obtained by using:
\begin{equation}
    \{p_x, p_y \} = \Big\{ \frac{m_{10}}{m_{00}}, \frac{m_{01}}{m_{00}} \Big\}    
    \label{eq:center_moments}
\end{equation}

To reduce the potential wobbling effects due to noise and the irregular folds appearing on the lumen, a moving average filter was applied considering the last four detected points in the sample window.
The output $\vb*{\hat{p}} = (\hat{p_x}, \hat{p_y})$ of the filter, is sent to the control module.

\subsection{Control Scheme}

%
In order to autonomously navigate through the lumen, we propose a image-based visual servoing strategy based on an eye-in-hand robot set-up. The navigation task means advancing the robot through the lumen, while keeping it in the center region to avoid the tip of the robot colliding with the inner walls and produce unintentionally perforations. 
In terms of control, this implies two high level tasks: \begin {enumerate*}
\item aligning the robot pose respect to the detected center, and once this has been achieved within a certain radius $\delta_c$, \item moving the robot forward at a constant speed.
\end {enumerate*}

The aim of image-based visual servoing is to find a mapping relationship $g$ between the task space $\Omega^X$, defined in the image pixel plane, and the robot actuation space $\Omega^Q$. 
In this work, we define $\vb*{q}(k) \in \Omega^Q$ as the actuators input at update step $k$; $\vb*{s}(k) \in \Omega^S$ as the robot configuration under input $\vb*{x}(k)$ and $\vb*{x}(k) \in \Omega^X$ as the input in the task space.

Considering movements in small steps, and a constant time step $\Delta t$, the transitions from one configuration to another in the task space due to the input difference $\vb*{q}(k)$ can be defined as:
\begin{equation} 
    \Delta \vb*{x}(k) = g(\Delta \vb*{q}(k))
\end{equation}
where $\Delta \vb*{q}(k)=\vb*{q}(k+1)-\vb*{q}(k)$ is the difference between actuator inputs at update steps $k$ and $k+1$.

When the kinematic model is known, the Jacobian $J$ is used to obtain this relationship. 
Given the characteristics of the proposed flexible robotic arm, for which kinematic models are not as accurate as for robots with rigid links, we opted for a model-less approach. 
An initial approximation of the image Jacobian $\hat{J}$ can be obtained using: \begin{equation}
    \hat{J}= \begin{bmatrix} \frac{\Delta f(q)^T}{\Delta q_1} & ... & \frac{\Delta f(q)^T}{\Delta q_n} \end{bmatrix}
\end{equation}
where $q_n$ indicates the $n$ motor revolution, $f(q)$ is the position of the target in the image plane coordinate system and $n$ is the number of actuation variables.
The data used for approximating $\hat{J}$ is obtained from commanding the robot to actuate each cable individually in small steps, and the feature points $\vb*{x}(k)$ detected at its corresponding $\vb*{q}(k)$.
During the pose correction step, the robot is only bending and not moving forward. Therefore, for this step, the last column of $\hat{J}$ is replaced by zeros. Noted that considering the movement of the robot tip has only a small movement, the Jacobian matrix is not updated during the movement.

To actuate motors in charge of the lumen centering task, we implement a resolved rates approach, which could provide us with smoother movements desired for surgical applications, the objective is to generate a control signal for the actuators inputs velocity $\vb*{\dot{q}} \triangleq \Delta q / \Delta t$ in terms of $\va*{v}$. 
The relationship between $\vb*{\dot{q}}$ and $\va*{v}$ is defined as:
\begin{equation}
    \vb*{\dot{q}} = \hat{J^{+}} \frac{\Delta x}{\Delta t}\triangleq \hat{J^{+}} \va*{v}
\label{eq:velocity_realtonship}    
\end{equation}
where $J^{+}$ is the Moore-Penrose pseudo inverse of the estimated Jacobian, $\va*{v}$ is the velocity in the task space.

\begin{figure}[tbp]
    \centering
    \includegraphics[width=0.4\textwidth]{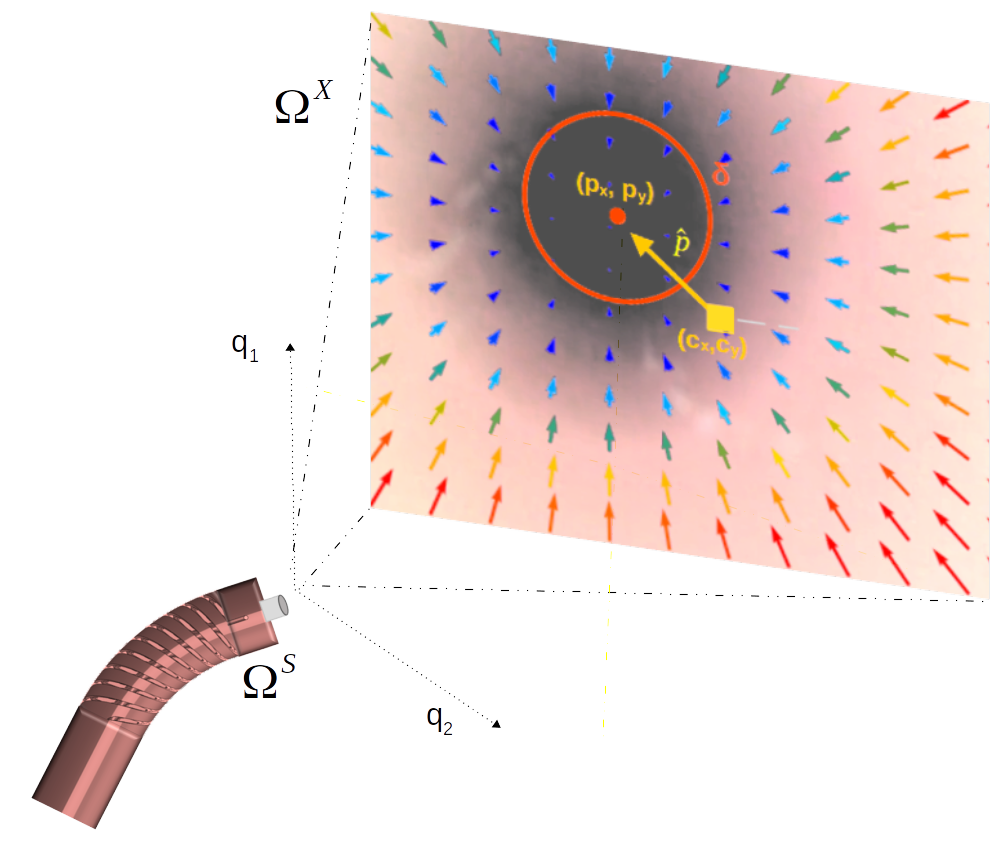}
    \caption{Diagram depicting the main idea of the \emph{Artificial Potential Well} approach. 
    The robot tries to adjust its configuration $\Omega^S$ by actuating $q_1$ and $q_2$ , to match the the center of the image plane $(c_x, c_y)$ to the detected center of the lumen $(p_x, p_y)$
    An overlay view from the endoscopic camera and the 2D representation of the potential well approach is depicted on the right. }
    \label{fig:potential_field_control}
\end{figure}

\begin{figure*}[tbp]
    \centering
    \includegraphics[width=0.80\textwidth]{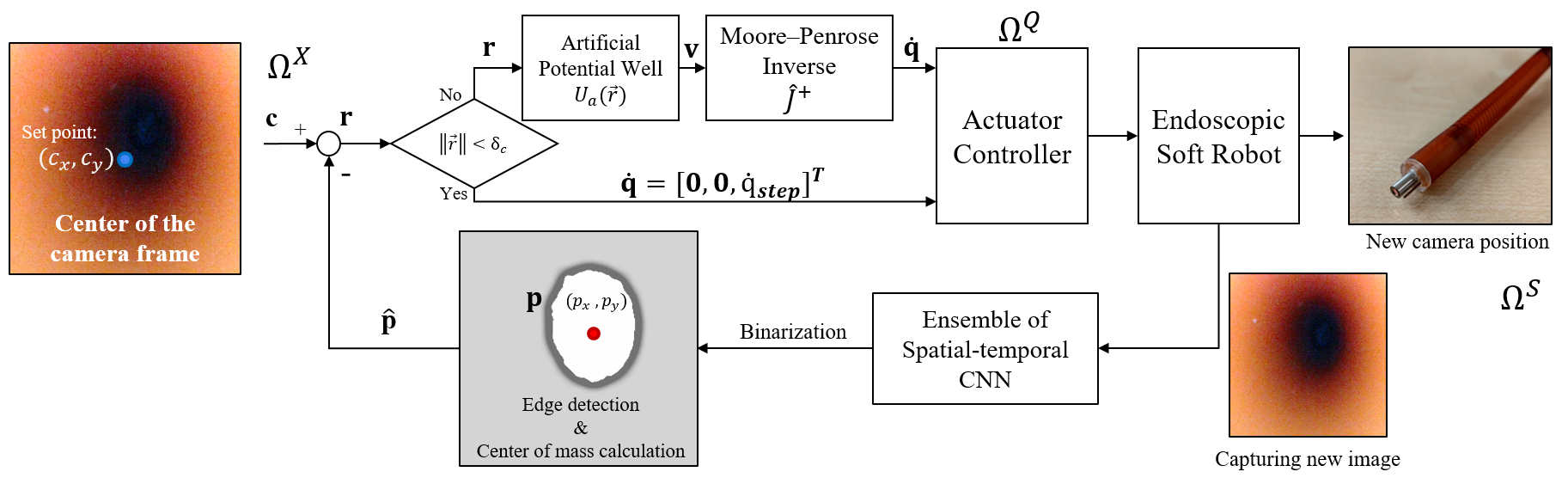}
    \caption{Control architecture of the proposed model-less visual servoing system. Given a target point $\vb*{p} = (p_x, p_y)$, detected by the lumen center detection module, its position is compared with set point $\vb*{c} = (c_x, c_y)$ to obtain the error $\vb*{r}$.
    The velocity $\vb*{v}$ is determined using the \emph{artificial potential well} according to the obtained $\vb*{r}$, which is translated to the motors $\vb*{\dot{q}}$ using the Moore-Penrose inverse $\hat{J^{+}}$. 
    If the norm of $\vb*{r}$ is below a threshold value $\delta_c$ the robot will move forward.
    At every step the robot position is updated and the endoscopic camera captures new images. The new image will first be stacked feedback to the CNN to segment the lumen image into binary image and detect the subsequent target point.
    }
    
    \label{fig:low_level_control_data_flow2}
\end{figure*}

The desired behaviour for $\va*{v}$ would be that when closer to the target, the smoother the movement is, whereas further away from its objective, the movement would be faster but up to a certain limit. 
Having these considerations in mind, we proposed a method in which the velocity commanded to the robot has a direct non-linear correspondence between the task space $\Omega^X$ defined in the image plane, and the velocity actuation space. 
The way of modeling this behaviour is by proposing an additional mapping. 
In this case we opted to implement an \emph{Artificial Potential Well} $U_a(\Vec{r})$, designed to perform an attractive action between the detected center of the lumen $\vb*{p} = (p_x, p_y)$ and the Set Point (SP), the center of the image plane $\vb*{c} = (c_x, c_y)$, with error $\va*{r}$ defined as the vector between $\vb*{p}$ and $\vb*{c}$.
The attraction action presents a linear behaviour in most of the space, except in the region close to the center target $\rho<\delta$, where a quadratic-behaviour potential is proposed in order to avoid singularities.
$\rho$ is defined as the norm of $\va*{r}$, and $\delta$ is the designed border:
\begin{equation}
    U_{a}(\va*{r}) = 
    \begin{cases}
        \frac{1}{2} \psi_1 \left|\left|\va*{r}\right|\right|^2  &;  \rho < \delta \\ 
        \psi_2 \left|\left|\va*{r}\right|\right| + \kappa &;  \rho \geq \delta
    \end{cases}
\label{eq:potential_field}
\end{equation}
$\psi_1$ is a proportionality constant defined as $\psi_1 = min \left[ 1, \rho/\delta \right]$, $\psi_2 =\delta \psi_1$ and $\kappa$ is a constant to ensure continuity in the boundary $\rho=\delta$.
A graphical representation of the $U_a(\Vec{r})$ is presented in Fig.~\ref{fig:potential_field_control}.
The relationship to link the potential well with the robot velocity is given by:  
\begin{equation}
    \nabla U_a = m \frac{d\va*{v}}{dt}
\label{eq:potential_force}    
\end{equation}
From which $\va*{v}$ can be obtained by integrating Eq.~(\ref{eq:potential_force}) and substituting  in Eq.~(\ref{eq:velocity_realtonship}) to obtain the values of $\vb*{\dot{q}}$ in the actuation space $\Omega^X$. 
In this case $m$ is just considered a proportionality constant which is set to the unit for simplicity.  
The values of $\vb*{\dot{q}}$ are sent to the actuator controller which contains two PID and one proportional controller for two DC motors and for the linear stage, respectively. 
Note that the forward movement is only allowed when $\rho < \delta_c$ in which case, the insertion speed is set to a constant value $\dot{q}_{step}$, thus $\vb*{\dot{q}} = [0, 0, \dot{q}_{step}]^{T}$.
A complete diagram depicting the complete control strategy in this section is shown in Fig.~\ref{fig:low_level_control_data_flow2}.

\section{System Validation}
\label{section:validation}
Ten ureter phantoms of different colors and diameters were manufactured using silicone-based liquid polymer, Dragon Skin (Smooth-On Inc.).
The phantoms have a tubular shape and are easy to bend. A sample is depicted on Fig.~\ref{fig:robot_setup_up}. 

\begin{figure}[tbp]
    \centering
    \includegraphics[width=0.45\textwidth]{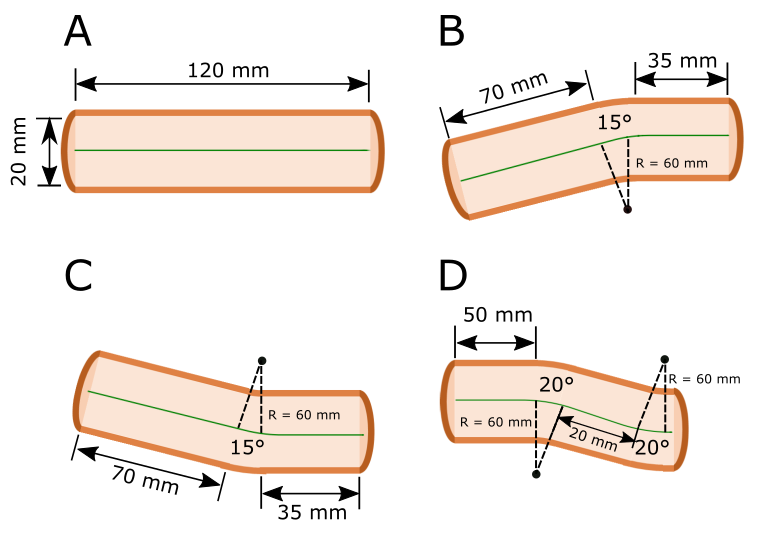}
    \caption{Pathways considering for testing the navigation task of the robot: 
    $A$) Straight line;  
    $B$) Left curve; 
    $C$) Right curve;  
    $D$) Two continuous curves.
    }
    \label{fig:paths}
\end{figure}

To resemble the curved nature of real endoluminal organs, four different pathways were considered as depicted in Fig.~\ref{fig:paths}. 
To have a reproducible ground-truth path, four molds were 3D printed as the designed pathways and the phantom was placed inside them. 
On the tip of the robot, three EM sensors were installed at an equidistant radius from the center and in an equilateral triangle configuration. 
The position of the robot tip was monitored using an EM tracking Aurora Planar 20-20 system (Northern Digital Inc, Canada). 
The EM field generator was set next to the robot tip and the experimental set-up as shown in Fig.~\ref{fig:robot_setup_up}.
To validate the modules in the proposed endoscopic system, different set of experiments for \emph{Lumen Segmentation}, \emph{Robot Centering} and \emph{Autonomous Intraluminal Navigation} were conducted. 
The lumen segmentation task was tested separately a priory before integrating it with the visual servoing module.  
The experimental protocols and the performance metrics for each task are described in detail in the following subsections.
\subsection{Lumen Segmentation Task}
Using the endoscopic camera (MC2, Redlemon), different from the one installed in the robotic model, video-clips from the inside of each of the phantoms were recorded, and frames were extracted to generate the datasets for training and validate the lumen segmentation module. 
In total 3,387 frames were used for training and validation of the network. 1,719 of this frames were extracted from the phantom video-clips whereas 1,668 frames came from videos of 4 patients undergoing ureteroscopy.          
A case-wise hold-out strategy was used to test the performance of the network. 
A total of 277 frames were used as test dataset and these frames were obtained from videos of the phantom used as well in the autonomous navigation experiments. The phantom used during test was manufactured using a different hue than those used for training. 
A 3-fold cross validation strategy was used to determine the CNN optimal hyperparameters: learning rate and batch size. 
The metric used to determine the best model was the Dice Coefficient ($DSC$) defined as $DSC = 1 - L_{DSC}$.

Once the hyper-parameters were chosen, the network was retrained randomly splitting the dataset in a ratio 70/30 regarding training and validation data. 
The test dataset corresponds to 277 image frames from two phantoms which were hold out from any previous training data. 
The images were manually labeled by 3 independent experts and the final ground truth was defined as the intersection areas proposed by each of them. 
An ablation study was performed comparing each of the separate branches $b_1$, which processed only Single Frames (SF), and $b_2$, which processed consecutive multiple frames (MF), against the proposed network consisting of the ensemble of SF and MF. 

The average center detection time is 0.09s deployed on a NVIDIA GeForce RTX 2080 GPU, using Python 3.5 and Tensorflow 2.4. 

\subsection{Robot Centering Task}
The purpose of the robot centering task is to test the performance and response of the proposed visual servoing architecture. 
For this task, the pathway $A$ with straight profile was used and the robot tip was placed at the opening of the phantom. 
The initial orientations of the tip were set manually by commanding the robot to point beyond a radial distance of 320 pixels from the center, detected by the center detection algorithm. 
Ten experiments with random initial orientations were carried out.
Common specifications, e.g. Steady-State Error (SSE), Rising Time (RT), Settling Time (ST), and Over-Shooting (OS) were used as performance metrics. 

\begin{figure}[tbp]
    \centering
    \includegraphics[width = 0.48\textwidth]{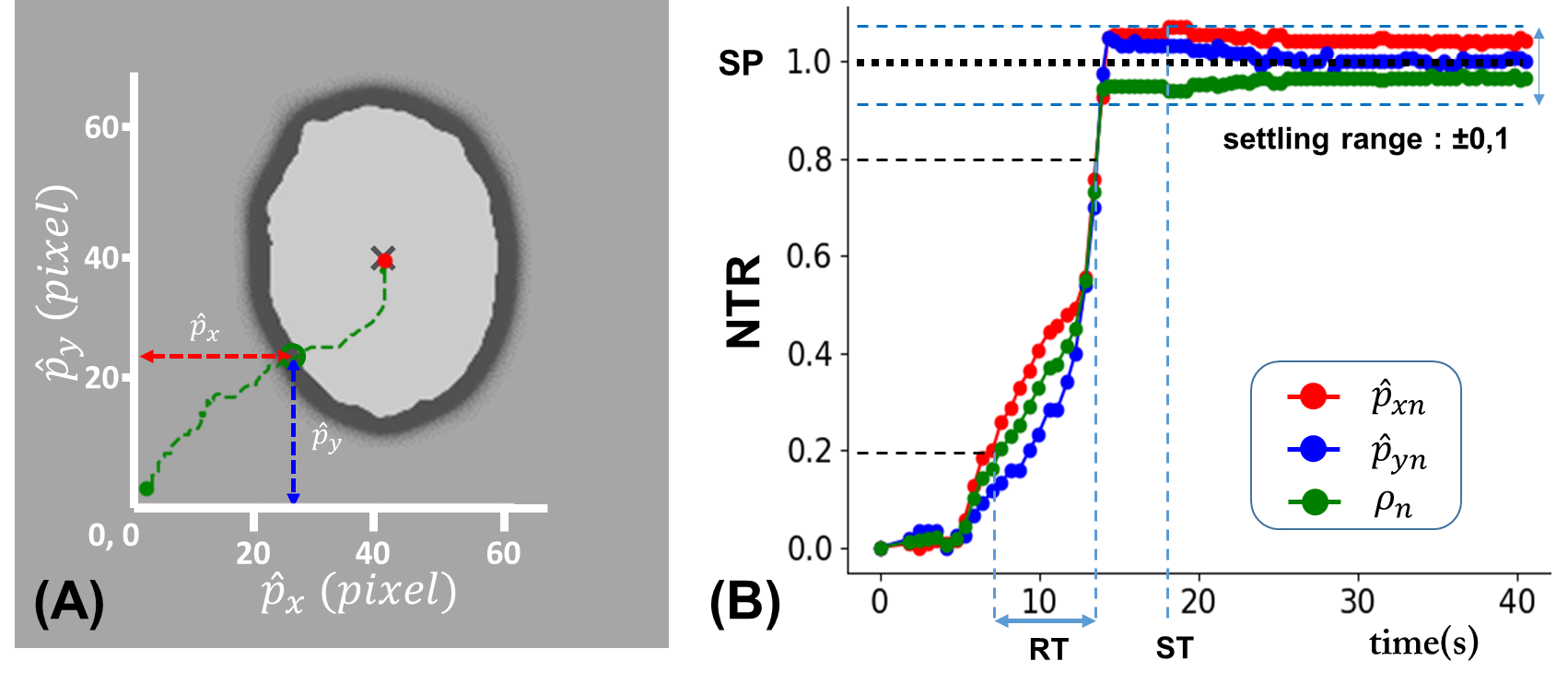}
    \caption{
    Response of the Potential well controller to the detected center of the lumen. 
    a) Sample of detected target, and the trajectory followed by the robot.
    The cross represents the detected target $\vb*{\hat{p}} = (\hat{p}_x, \hat{p}_y)$ at the beginning of the motion $t=0$.  
    The dashed green line is the trajectory followed by the robot and the green circle on the trajectory is $\vb*{\hat{p}}$ on any time $t=0$.   
    b) Normalized Target Response values, $\hat{p}_{xn}(t), \hat{p}_{yn}(t)$ and $\rho_{n}(t)$, of the robot centering task.
    The Settling Time is the last value of $\rho_n(t)$ that falls within a threshold of $\pm$0.1 respect to the set point. 
    The rising time is the time for the normalized target distance  $\rho_n(t)$ to rise from 0.2 to 0.8.
    }
    \label{fig:robot_resp}
\end{figure}

In order to compare results from different experiments, we define three Normalized Target Response values (NTRs), $\hat{p}_{xn}(t), \hat{p}_{yn}(t)$ and $\rho_{n}(t)$. 
The subscript $n$ indicates that $\hat{p}_{xn}(t)$, $\hat{p_{y}n}(t)$ and $\rho_{n}(t)$ are normalized from the recorded target point $\hat{p}_x(t)$, $\hat{p}_y(t)$ and target distance $\rho(t)$ over time stamp $t = t_1, t_2, ... t_i$, respectively. The three NTRs are defined as:

\begin{equation}
\label{eq:normalized_coordinate_x_y}
    \{ \hat{p}_{xn}(t_{i}) , \hat{p}_{yn}(t_{i})\} = \Big\{ \frac{\hat{p}_{x}(t_{0}) - \hat{p}_{x}(t_{i})}{\hat{p}_{x}(t_{0})},
    \frac{\hat{p}_{y}(t_{0}) - \hat{p}_{y}(t_{i})}{\hat{p}_{y}(t_{0})} \Big\}
\end{equation}
\begin{equation}
\label{eq:normalized_distance}
    \rho_{n}(t_{i}) = \frac{\rho(t_{0}) - \rho(t_{i})}{\rho(t_{0})}
\end{equation}

where $t_{0}$ is the initial time when the experiment starts and $t_{i}$ is any time stamp in $t$. Each NTR in each experiment starts with a value 0.0 at $t_{0}$. When the target point is reached, the NTR has a value 1.0, which is defined as the Set Point(SP) of the response of each experiment. With the NTRs, we can define the performance metrics for this task as follows:

\begin{itemize}
    \item SSE: The percentage error from the SP to the $\rho_n(t_{i})$  when the robot stops moving.
    \item RT: The time for $\rho_n(t)$ to rise from 0.2 to 0.8
    \item ST: The time when the last value of $\rho_n(t)$ that falls to within $\pm$0.1  from the SP
    \item OS: This is defined in each of the coordinates $x$ and $y$ axes, instead of the distance. 
    The over-shooting at each coordinate is the percentage error between the maximum $\hat{p}_{xn}(t)$ and $\hat{p}_{yn}(t)$ and the SP, in case $\hat{p}_{xn}(t)$ or $\hat{p}_{yn}(t)$ is larger than the SP. 
\end{itemize}
An example of the response curves from one of the experiments is shown in Fig.~\ref{fig:robot_resp}.

\subsection{Autonomous Intraluminal Navigation Task}
In the intraluminal navigation task, the proposed endoscopic robot system should autonomously navigate through the lumen in all the paths defined in Fig.~\ref{fig:paths}. At each experiment the starting point was defined at the opening of the phantom with a random orientation and the insertion direction is always vertical to the opening of the lumen. 
The goal line was set as a virtual crossing line, perpendicular to the displacement axis of the linear stage in the EM reference frame. 
The traveling distance along the axis of the linear stage is on average 130 mm long.   
The values of $\delta$ and $\delta_c$, which allow the movement forward, were set empirically to 25 and $0.6 \times \delta$ respectively.
Five experiments were carried out with path $A$, and for each path $B$, $C$ and $D$, 15 experiments were carried out for each of them. 
Calibration of the EM tracking system was done before each experiment by measuring the eight corners of the 3D-printed molds of which has known dimensions. 
The orientation of the theoretical ground-truth path was then registered to the EM tracking system reference frame using Iterative Closest Points algorithm. 

Errors and smoothness of the trajectory were considered as the performance metrics for the autonomous intraluminal navigation task and they are defined as follows:

\begin{itemize}
    \item Completion Time (CT): The time when the proposed robot system completes the task.
    \item Mean Absolute Error (MAE): The absolute error $e_z$ was compared between the ground-truth path and the measurement of EM tracking sensors along the depth direction. The MAE is the sum of $e_z$ divided by the number of data points.
    \item Max Absolute Error (MaxAE): Largest $e_z$ along the depth axis.
    \item Log-Dimensionless Jerk (LDJ): Is the negative value of the natural logarithm of the mean absolute jerk, normalized by the peak speed and multiplied by the trial duration~\cite{gulde2018smoothness}, defined as:
    \begin{equation}
        LDJ = -ln \left( \frac{\Delta t}{v_p^2} \int^{t_{f}}_{t_i} \bigl\lvert \frac{dv^2}{dt^2}\bigl\lvert^2  dt\right)
    \end{equation}
    where $\Delta t$ is the trial duration and $v_p$ is the peak speed. 
    \item Spectral arc-length (SPARC): As defined in ~\cite{balasubramanian2011robust}, it refers to a smoothness metric which measures the arc length of the Fourier magnitude spectrum of the speed profile $v(t)$ within an adaptive frequency range. 
    \begin{equation}
        SPARC = -\int^{\omega_c}_{0} \sqrt{\left( \frac{1}{\omega_c}\right)^2 +\left( \frac{d \Hat{V}(\omega)}{d \omega}\right)^2 }d\omega
    \end{equation}
    where $V(\omega)$ is the Fourier magnitude spectrum of $v(t)$, and $\left[0, \omega_c \right]$ is the frequency band occupied by the given movement. $\Hat{V(\omega)}$ is the normalized amplitude spectrum. 
    \item Number of peaks (NP): defined as the number of velocity profile peaks exceeding a prominence of 0.05 respect to its neighbors divided by the traveled path length
\end{itemize}
By definition LDJ and SPARC should have negative values and results closer to zero represents smoother movements.  

\section{Results and Discussion}
\label{section:results}
\subsection{Lumen Segmentation Task}
\begin{figure}[tbp]
    \centering
    \includegraphics[width=0.33\textwidth]{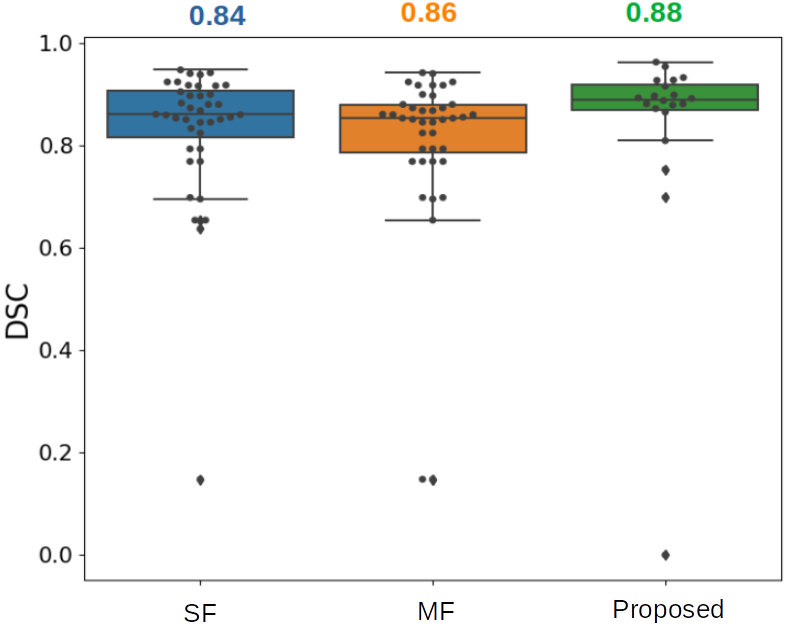}
    \caption{Boxplots of the $DSC$ values for the different lumen segmentation networks tested. 
    SF:~Single input-frame network (branch $b_1$), 
    MF:~3 consecutive frames (branch $(b_2$) and the Proposed network consisting of the ensemble of SF and MF networks.}
    \label{fig:DSC_boxplots}
\end{figure}
The median $DSC$ values obtained on the test dataset for each of the networks were 0.84, 0.86 and 0.88, for SF, MF, and the proposed network respectively. Figure~\ref{fig:DSC_boxplots} shows the results from these experiments. 
The Kruskal-Wallis test was used to determine statistical significance among the models, however, no statistical significance was found. This might be related to the fact that the dataset in which it was tested does not contain challenging cases.   
The ensemble model, which obtained the best performances and in previous work has shown to be the more robust against conditions variability and artifacts~\cite{lazo2021using}, was chosen to be implemented in the visual servoing module. 

\subsection{Robot Centering Task}
\begin{table}[bp]
    \centering    
    \caption{Results of the Robot Centering Task.}
    \begin{tabular}{cc}
    Metric & Avg $\pm$ STD \\ 
    \hline
    Steady state error(\%) &  5.84 $\pm$ 2.67 \\
    Rising time(s)         &  8.34 $\pm$ 1.16 \\
    Settling time(s)       &  27.0 $\pm$ 15.9 \\
    Over-shooting in x(\%) &  11.8 $\pm$ 7.58 \\
    Over-shooting in y(\%) &  4.02 $\pm$ 3.59 \\
    \hline
    \end{tabular}
    \label{table:result_task_1}
\end{table}
The results for the robot centering task are presented in Table \ref{table:result_task_1}. 
The robot was able to reduce the error bellow a 10\% from target distance, except for one case (SSE = 11\%). Most of the trials reached the SSE within 25 seconds and two trials needed almost 50 seconds to settle. 
It is likely that in those trials with higher ST or bigger error, the targets fell into the dead-zone of the robot. 
This issue for this cable-driven mechanism should be included in the future work to further improve the performance of the SSE.
There is a noticeable difference of the average OS concerning the $x$-axis and the $y$-axis. 
On the $y$-axis, no trial exceeded more than 5\% OS. On the other hand, for the $x$-axis, three trials exceeded 10\% and one exceeded 20\%. 
This might be caused by the weight of the robot tip and could possibly be solved by providing a non-radial-homogeneous potential well in future work.
The results from this task clearly showed that the proposed robot is able to correct itself from a random starting pose given a visible target. 
Despite the slow response, in all the cases the robot is able to center the camera in the desirable target region. 

\subsection{Autonomous Intraluminal Navigation Task}

\begin{figure*}[tbp]
    \centering
    \includegraphics[width=0.99\textwidth]{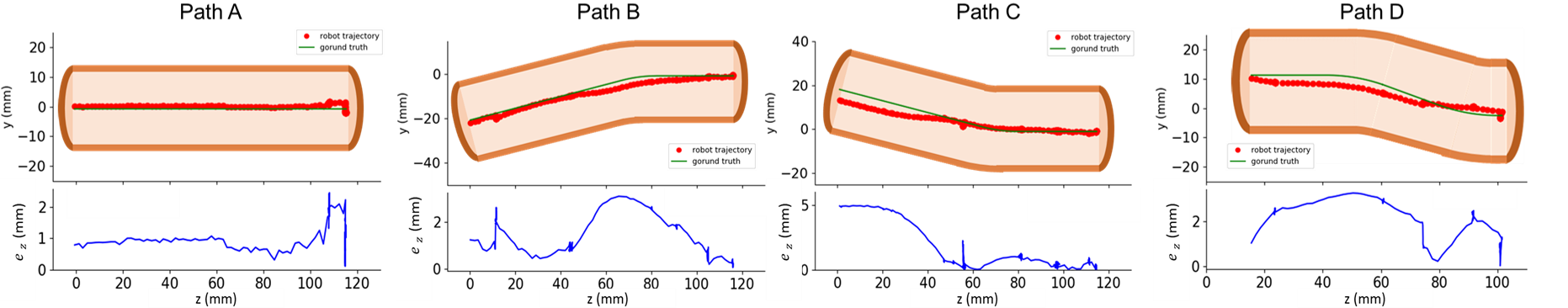}
    \caption{Sample results of the autonomous intraluminal navigation for each of the four different paths. 
    The top figures show the comparison between the path followed by the robot and the ground truth path in the center of the lumen phantom. 
    The bottom plots show the absolute error $e_z$ between the ground truth path and the robot tip position measured by the EM tracking sensor along the path axis.}
    \label{fig:sample_results_navigation}
\end{figure*}

For each path 15 experimental trials were performed. 
Figure~\ref{fig:sample_results_navigation} shows examples of the path followed in each of the different scenarios along with the respective absolute error graph. 

In all the cases, the robot was able to complete the task by reaching the goal point. 
The average CT is the shortest in path $A$ with a time of 81.2$\pm$7.28 s, and the longest in path $D$ with 212.9 $\pm$57.8 s. 
The results are as expected given that the geometry in path $A$ is a straight way while the path $D$ has a S-shaped curve which is considered more difficult. 
For paths $B$ and $C$, which are symmetrical across the displacement axis, there is a difference on CT of 37.8 s.


The highest MAE was in path $B$ with 2.17$\pm$0.34 mm and was the lowest for path $A$ with 0.86$\pm$0.33 mm. For MaxAE path $B$ presents the highest value again with 6.11$\pm$0.37 mm and path $A$ presents the lowest value with 2.09$\pm$0.23 mm. 
The difference in these metrics between paths $B$ and $C$ could be related to manufacturing issues and asymmetrical elongation of the tendon wires after repetitive tension. 
However, it was observed that in all cases, the robot stopped moving forward several times and corrected its orientation avoiding collisions with the inner wall.  

Regarding smoothness there is no significant difference between the different paths in terms of LDJ and SPARC.
In path $A$ the lowest values are obtained for both metrics with 12.31$\pm$0.40 and -9.16$\pm$0.15 respectively, and the highest values are on path $D$ with  15.3$\pm$1.10 and -9.80$\pm$0.34. 
As for NP the behaviour is similar, the difference between the best ($A$) and the worst ($D$) performance is of 63.9. 
These results are understandable since the regularity and precision of the movement of the robot is steady regardless the testing path.

Path $D$ corresponds to a more realistic and complex scenario where the lumen twists in consecutive curvatures. 
Having in mind that the inner diameter of the lumen is 15 mm, and the path followed by the robot presents an average error bellow 2 mm and a maximum error of 4.35 mm, this indicates the robot is following the center-line of the phantom lumen and avoiding collisions with the inner walls. This also implies that implementation of the endoscopic robot using vision feedback instead of position feedback seems plausible for autonomous navigation in narrow lumen.
Nevertheless, the average CT is around 3.5 minutes for an average traveling distance of 130 mm, which needs to be improved.  

Also it was observed that sometimes the correction of the trajectory happened at a very early stage, when a curvature of the lumen was detected, but its actual position was further in the back, delaying the forward movement. 
This might be related to the current implementation which is not able to determine depth from video frames.   

\begin{table}[tbp]
\scriptsize
\caption{
\scriptsize
Results of the intraluminal navigation task for the 4 different paths.}
\centering
\begin{tabular}{ccccc}
     & Path $A$ & Path $B$ & Path $C$ & Path $D$ \\
\hline
Metrics*   & Avg $\pm$ STD  & Avg $\pm$ STD  & Avg $\pm$ STD  & Avg $\pm$ STD \\ 
\hline
CT (s)     & 81.2 $\pm$70  & 119.7$\pm$33 & 157.5$\pm$57 & 212.9$\pm$57 \\
MAE (mm)   & 0.86$\pm$0.33  & 2.17$\pm$0.34  & 1.74$\pm$0.32  & 1.99$\pm$0.29  \\
MaxAE (mm) & 2.09$\pm$0.23  & 6.11$\pm$0.37  & 5.12$\pm$0.56  & 4.35$\pm$0.41  \\
LDJ        & -12.5$\pm$0.40 & -13.5$\pm$1.42 & -14.4$\pm$1.41 & -15.3$\pm$1.10 \\
SPARC      & -9.16$\pm$0.15 & -9.16$\pm$0.27 & -9.48$\pm$0.37 & -9.80$\pm$0.34 \\ 
NP         & 105.2$\pm$14 & 130.8$\pm$42     & 163.06$\pm$58  & 169.10$\pm$40 \\ \hline
\multicolumn{5}{l}{\scriptsize{*Metrics: Completion Time (CT), Mean Absolute Error (MAE), Max Absolute}}\\
\multicolumn{5}{l}{\scriptsize{Error (MaxAE), Log-dimensionless Jerk (LDJ), Spectral Arc Length (SPARC)}}\\
\multicolumn{5}{l}{\scriptsize{and number of peaks (NP).}}
\end{tabular}%
\end{table}

\section{Conclusion}

In this work we presented a 3D printed flexible robotic endoscope integrated with a model-less visual servoing method, based on CNNs, for autonomous intraluminal navigation. 
The results obtained show that the robot is able to find the center of the lumen and correct its position to safely navigate through different pathways not previously seen by the robot. 


With this study we showed that it is possible to use a CNN previously trained in real-patient data, and adapt it to make it work for visual servoing in phantoms. 
Validation in more realistic scenarios is considered as a future task, as well as a comprehensive comparison with human steering manually, semi-automatically or guided. 

Further work includes finding alternatives for depth estimation from monocular images, which could improve navigation, and the integration of computer vision modules for tissue segmentation that could require a more detailed examination.  

The results obtained in this work show that automation of certain tasks in endoscopic interventions is possible, and opens the way towards further development of robotic models and new control strategies to aid in endoscopic interventions.  

\bibliographystyle{IEEEtran}
\bibliography{bibliography.bib}

\begin{thebibliography}{10}
\providecommand{\url}[1]{#1}
\csname url@samestyle\endcsname
\providecommand{\newblock}{\relax}
\providecommand{\bibinfo}[2]{#2}
\providecommand{\BIBentrySTDinterwordspacing}{\spaceskip=0pt\relax}
\providecommand{\BIBentryALTinterwordstretchfactor}{4}
\providecommand{\BIBentryALTinterwordspacing}{\spaceskip=\fontdimen2\font plus
\BIBentryALTinterwordstretchfactor\fontdimen3\font minus
  \fontdimen4\font\relax}
\providecommand{\BIBforeignlanguage}[2]{{%
\expandafter\ifx\csname l@#1\endcsname\relax
\typeout{** WARNING: IEEEtran.bst: No hyphenation pattern has been}%
\typeout{** loaded for the language `#1'. Using the pattern for}%
\typeout{** the default language instead.}%
\else
\language=\csname l@#1\endcsname
\fi
#2}}
\providecommand{\BIBdecl}{\relax}
\BIBdecl

\bibitem{fisher2011complications}
D.~A. Fisher, J.~T. Maple, T.~Ben-Menachem, B.~D. Cash, G.~A. Decker, D.~S.
  Early, J.~A. Evans, R.~D. Fanelli, N.~Fukami, J.~H. Hwang \emph{et~al.},
  ``Complications of colonoscopy,'' \emph{Gastrointestinal Endoscopy}, 2011.

\bibitem{da2020challenges}
T.~da~Veiga, J.~H. Chandler, P.~Lloyd, G.~Pittiglio, N.~J. Wilkinson, A.~K.
  Hoshiar, R.~A. Harris, and P.~Valdastri, ``Challenges of continuum robots in
  clinical context: A review,'' \emph{Progress in Biomedical Engineering},
  vol.~2, no.~3, 2020.

\bibitem{ali2020objective}
S.~Ali, F.~Zhou, B.~Braden, A.~Bailey, S.~Yang, G.~Cheng, P.~Zhang, X.~Li,
  M.~Kayser, R.~D. Soberanis-Mukul \emph{et~al.}, ``An objective comparison of
  detection and segmentation algorithms for artefacts in clinical endoscopy,''
  \emph{Scientific Reports}, vol.~10, no.~1, 2020.

\bibitem{dankelman2011current}
J.~Dankelman, J.~J. Van Den~Dobbelsteen, and P.~Breedveld, ``Current technology
  on minimally invasive surgery and interventional techniques,'' in \emph{2011
  2nd International Conference on Instrumentation Control and
  Automation}.\hskip 1em plus 0.5em minus 0.4em\relax IEEE, 2011, pp. 12--15.

\bibitem{de2006handling}
J.~J. de~la Rosette, T.~Skrekas, and J.~W. Segura, ``Handling and prevention of
  complications in stone basketing,'' \emph{European urology}, vol.~50, no.~5,
  pp. 991--999, 2006.

\bibitem{bergeles2013passive}
C.~Bergeles and G.-Z. Yang, ``From passive tool holders to microsurgeons:
  safer, smaller, smarter surgical robots,'' \emph{IEEE Transactions on
  Biomedical Engineering}, vol.~61, no.~5, 2013.

\bibitem{attanasio2021autonomy}
A.~Attanasio, B.~Scaglioni, E.~De~Momi, P.~Fiorini, and P.~Valdastri,
  ``Autonomy in surgical robotics,'' \emph{Annual Review of Control, Robotics,
  and Autonomous Systems}, vol.~4, pp. 651--679, 2021.

\bibitem{boehler2020realiti}
Q.~Boehler, D.~S. Gage, P.~Hofmann, A.~Gehring, C.~Chautems, D.~R. Spahn,
  P.~Biro, and B.~J. Nelson, ``Realiti: A robotic endoscope automated via
  laryngeal imaging for tracheal intubation,'' \emph{IEEE Transactions on
  Medical Robotics and Bionics}, vol.~2, no.~2, pp. 157--164, 2020.

\bibitem{wang2016visual}
H.~Wang, B.~Yang, Y.~Liu, W.~Chen, X.~Liang, and R.~Pfeifer, ``Visual servoing
  of soft robot manipulator in constrained environments with an adaptive
  controller,'' \emph{IEEE/ASME Transactions on Mechatronics}, 2016.

\bibitem{lai2020toward}
J.~Lai, K.~Huang, B.~Lu, and H.~K. Chu, ``Towards vision-based adaptive
  configuring of a bidirectional two-segment soft continuum manipulator,'' in
  \emph{2020 IEEE/ASME International Conference on Advanced Intelligent
  Mechatronics (AIM)}.\hskip 1em plus 0.5em minus 0.4em\relax IEEE, 2020.

\bibitem{wu2015model}
K.~Wu, L.~Wu, C.~M. Lim, and H.~Ren, ``Model-free image guidance for
  intelligent tubular robots with pre-clinical feasibility study: Towards
  minimally invasive trans-orifice surgery,'' in \emph{2015 IEEE International
  Conference on Information and Automation}.\hskip 1em plus 0.5em minus
  0.4em\relax IEEE, 2015.

\bibitem{girerd2020automatic}
C.~Girerd, A.~V. Kudryavtsev, P.~Rougeot, P.~Renaud, K.~Rabenorosoa, and
  B.~Tamadazte, ``Automatic tip-steering of concentric tube robots in the
  trachea based on visual slam,'' \emph{IEEE Transactions on Medical Robotics
  and Bionics}, vol.~2, no.~4, 2020.

\bibitem{fang2019vision}
G.~Fang, X.~Wang, K.~Wang, K.-H. Lee, J.~D. Ho, H.-C. Fu, D.~K.~C. Fu, and
  K.-W. Kwok, ``Vision-based online learning kinematic control for soft robots
  using local gaussian process regression,'' \emph{IEEE Robotics and Automation
  Letters}, vol.~4, no.~2, 2019.

\bibitem{wang2020eye}
X.~Wang, G.~Fang, K.~Wang, X.~Xie, K.-H. Lee, J.~D. Ho, W.~L. Tang, J.~Lam, and
  K.-W. Kwok, ``Eye-in-hand visual servoing enhanced with sparse strain
  measurement for soft continuum robots,'' \emph{IEEE Robotics and Automation
  Letters}, vol.~5, no.~2, 2020.

\bibitem{zhang2020enabling}
Q.~Zhang, J.~M. Prendergast, G.~A. Formosa, M.~J. Fulton, and M.~E. Rentschler,
  ``Enabling autonomous colonoscopy intervention using a robotic endoscope
  platform,'' \emph{IEEE Transactions on Biomedical Engineering}, 2020.

\bibitem{martin2020enabling}
J.~W. Martin, B.~Scaglioni, J.~C. Norton, V.~Subramanian, A.~Arezzo, K.~L.
  Obstein, and P.~Valdastri, ``Enabling the future of colonoscopy with
  intelligent and autonomous magnetic manipulation,'' \emph{Nature Machine
  Intelligence}, vol.~2, no.~10, 2020.

\bibitem{yang2017medical}
G.-Z. Yang, J.~Cambias, K.~Cleary, E.~Daimler, J.~Drake, P.~E. Dupont, N.~Hata,
  P.~Kazanzides, S.~Martel, R.~V. Patel \emph{et~al.}, ``Medical
  robotics—regulatory, ethical, and legal considerations for increasing
  levels of autonomy,'' \emph{Science Robotics}, vol.~2, no.~4, 2017.

\bibitem{prendergast2020real}
J.~M. Prendergast, G.~A. Formosa, M.~J. Fulton, C.~R. Heckman, and M.~E.
  Rentschler, ``A real-time state dependent region estimator for autonomous
  endoscope navigation,'' \emph{IEEE Transactions on Robotics}, vol.~37, no.~3,
  pp. 918--934, 2020.

\bibitem{lazo2021using}
J.~F. Lazo, A.~Marzullo, S.~Moccia, M.~Catellani, B.~Rosa, M.~de~Mathelin, and
  E.~De~Momi, ``Using spatial-temporal ensembles of convolutional neural
  networks for lumen segmentation in ureteroscopy,'' \emph{International
  Journal of Computer Assisted Radiology and Surgery}, 2021.

\bibitem{culmone2020exploring}
C.~Culmone, P.~W. Henselmans, R.~I. van Starkenburg, and P.~Breedveld,
  ``Exploring non-assembly 3d printing for novel compliant surgical devices,''
  \emph{Plos One}, vol.~15, no.~5, p. e0232952, 2020.

\bibitem{gulde2018smoothness}
P.~Gulde and J.~Hermsd{\"o}rfer, ``Smoothness metrics in complex movement
  tasks,'' \emph{Frontiers in Neurology}, vol.~9, 2018.

\bibitem{balasubramanian2011robust}
S.~Balasubramanian, A.~Melendez-Calderon, and E.~Burdet, ``A robust and
  sensitive metric for quantifying movement smoothness,'' \emph{IEEE
  Transactions on Biomedical Engineering}, vol.~59, no.~8, 2011.

\end{thebibliography}
%


\end{document}